\documentclass[10pt,twocolumn,letterpaper]{article}

\usepackage{cvpr}              

\usepackage{algorithm}
\usepackage{algorithmic}
\usepackage{multirow}
\usepackage[inline]{enumitem}
\usepackage{enumitem}
\usepackage{boldline}
\usepackage{hhline}
\usepackage{amsmath}

\definecolor{cvprblue}{rgb}{0.21,0.49,0.74}
\usepackage[pagebackref,breaklinks,colorlinks,allcolors=cvprblue]{hyperref}

\def\paperID{10941} 
\def\confName{CVPR}
\def\confYear{2025}

\title{\texttt{CL-MoE}: Enhancing Multimodal Large Language Model with Dual Momentum Mixture-of-Experts for Continual Visual Question Answering}
\author {
    Tianyu Huai\textsuperscript{\rm 1},
    Jie Zhou\textsuperscript{\rm 1}\thanks{Corresponding author, jzhou@cs.ecnu.edu.cn.} , 
    Xingjiao Wu\textsuperscript{\rm 1}, 
    Qin Chen\textsuperscript{\rm 1}, 
    Qingchun Bai\textsuperscript{\rm 2}, 
    Ze Zhou\textsuperscript{\rm 3}, 
    Liang He\textsuperscript{\rm 1} \\
    \textsuperscript{\rm 1}School of Computer Science and Technology, East China Normal University \\
    \textsuperscript{\rm 2}Shanghai Open University, Shanghai, China 
    \textsuperscript{\rm 3}ZhuQingTing Data Technology (Zhejiang) Co., Ltd. 
}

\begin{document}
\maketitle

\begin{abstract}
Multimodal large language models (MLLMs) have garnered widespread attention from researchers due to their remarkable understanding and generation capabilities in visual language tasks (e.g., visual question answering). 
However, the rapid pace of knowledge updates in the real world makes offline training of MLLMs costly, and when faced with non-stationary data streams, MLLMs suffer from catastrophic forgetting during learning.  
In this paper, we propose an MLLMs-based dual momentum Mixture-of-Experts (\texttt{CL-MoE}) framework for continual visual question answering (VQA). 
We integrate MLLMs with continual learning to utilize the rich commonsense knowledge in LLMs.
We introduce a Dual-Router MoE (RMoE) strategy to select the global and local experts using task-level and instance-level routers, to robustly assign weights to the experts most appropriate for the task.
Then, we design a dynamic Momentum MoE (MMoE) to update the parameters of experts dynamically based on the relationships between the experts and tasks/instances, so that the model can absorb new knowledge while maintaining existing knowledge. 
The extensive experimental results indicate that our method achieves state-of-the-art performance on 10 VQA tasks, proving the effectiveness of our approach. 
\end{abstract}

\section{Introduction}
\label{sec:intro}

In recent years, multimodal large language models (MLLMs)~\cite{li2020oscar,akbari2021vatt,chen2022visualgpt,yan2023video,liu2024visual} have attracted widespread attention for their outstanding abilities of understanding and generating in visual language tasks. These models typically employ pre-training to acquire comprehensive knowledge and utilize fine-tuning synchronized with human instructions. The pre-training phase focuses on aligning visual and language modalities through extensive data and various techniques. During the fine-tuning phase, these aligned models use meticulously crafted instruction datasets to enhance their ability to follow human instructions.

MLLMs have demonstrated remarkable abilities in learning new tasks and knowledge by training on offline data. 
However, training MLLMs with data streams in an incremental learning setting can result in forgetting previously acquired knowledge, known as catastrophic forgetting~\cite{french1999catastrophic,kirkpatrick2017overcoming}. Combining new instructions with the original ones for multi-task training from scratch can address this issue. Nevertheless, it is impractical due to the high costs and the relentless influx of data in the real world. Hence, it is essential to explore ways to follow new human instructions and assimilate new knowledge while preserving the original knowledge of MLLM as much as possible.

\begin{figure}[t]
\centering
\includegraphics[width=1\columnwidth]{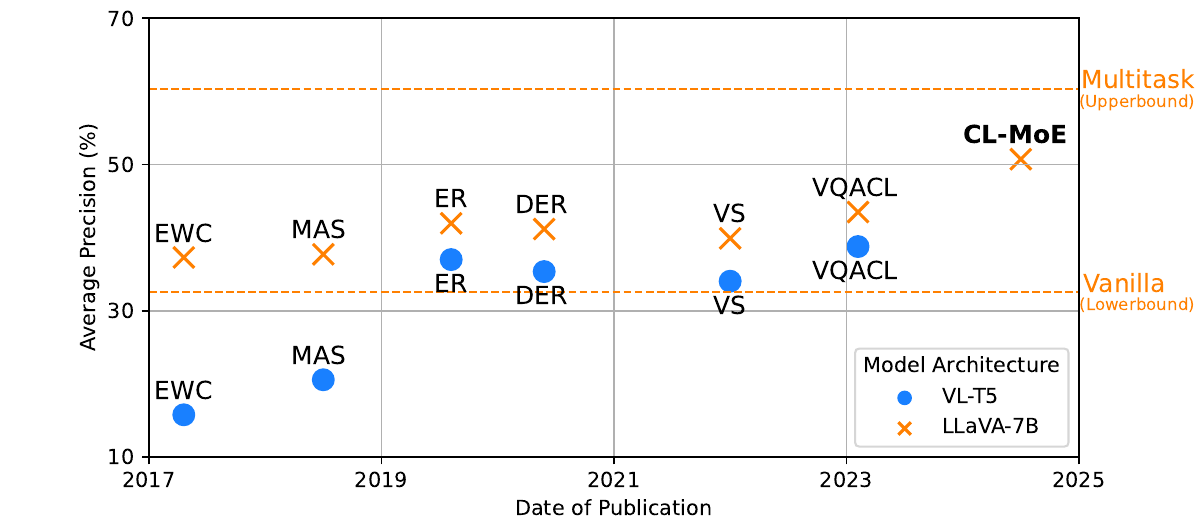}
\caption{Progress of continual learning over time on VQA v2. We give the results of previous CL methods based on \textcolor[RGB]{26,128,255}{VL-T5} and \textcolor[RGB]{242,133,0}{LLaVA}. Multitask represents the upper bound of the model, which trains over all the tasks once.}
\label{intro}
\end{figure}

Many efforts have been made previously about continual learning (CL) to improve catastrophic forgetting, which can be divided into regularization- and rehearsal-based methods~\cite{yang2024recent}.
For regularization-based methods, MAS~\cite{aljundi2018memory} estimates importance weights for network parameters in an unsupervised and online way, which allows the model to adapt to unlabeled data. NPC~\cite{paik2020overcoming} retains existing knowledge by controlling the plasticity of each neuron or filter in CNN and proposes a memory-efficient consolidation algorithm. 
For rehearsal-based methods, ER~\cite{chaudhry2019continual} proposes using a memory bank to store a tiny episodic memory to be trained jointly with the current task. DER~\cite{buzzega2020dark} proposes mixing rehearsal with knowledge distillation and regularization methods for general continual learning. 
Research on multimodal continual learning is also gradually emerging. VQACL~\cite{zhang2023vqacl} proposes a new VQA learning and reasoning paradigm and uses rehearsal methods to alleviate the forgetting problem. PROOF~\cite{zhou2023learning} addresses the challenge of catastrophic forgetting by freezing the image and text encoders and using expandable projections.

Despite the excellent effectiveness of the previous studies, there are several problems in the field of continual multimodal tasks.
First (\textbf{Q1}), multimodal tasks require reasoning based on rich commonsense knowledge about the world. As shown in Figure \ref{intro}, the models based on LLaVA~\cite{liu2024visual} outperform the corresponding ones based on VL-T5~\cite{cho2021unifying}.
Second (\textbf{Q2}), each task and instance may need multiple skills (experts), and each skill may serve more than one task or instance. Additionally, most cases can be solved with several fixed experts, and example-specific fine-grained experts can also improve performance. It is important to select the task-specific and instance-specific experts to generate the corresponding answer.
Third (\textbf{Q3}), through the preliminary experiments, we found that directly applying existing continual learning (CL) methods~\cite{kirkpatrick2017overcoming,aljundi2018memory,chaudhry2019continual,buzzega2020dark,wan2022continual,zhang2023vqacl} to MLLMs for visual-language tasks still have the problem of catastrophic forgetting. 
Compared with models that trained overall tasks at once (e.g., Multitask), training the models on the sequence of tasks one by one (e.g., Vanilla, VQACL) will result in a significant performance drop.

To address these issues, we propose a dual momentum Mixture-of-Experts (MoE) framework (\texttt{CL-MOE}) based on MLLMs for continual visual question answering. 
For \textbf{Q1}, we integrate continual visual question answering (VQA) with MLLMs to fully use the potential of MLLMs, which have outstanding reasoning abilities with rich world knowledge. 
For \textbf{Q2}, we design a Dual-Router MoE (RMoE), which consists of task-level and instance-level routers. In this way, our model captures appropriate experts from local and global perspectives by considering the task and instance at the same time.
For \textbf{Q3}, we introduce a dynamic Momentum MoE (MMoE) to update the parameters of experts dynamically based on the correlations between the experts and the tasks/instances using a momentum mechanism, assimilating new knowledge while mitigating catastrophe forgetting. 
The experiments show that our \texttt{CL-MOE} outperforms several strong baselines on the widely used VQA v2~\cite{goyal2017making} dataset. It indicates that our model effectively mitigates catastrophic forgetting and promotes the model's forward and backward transfer ability. Ablation studies also prove the effectiveness of each component in \texttt{CL-MoE}.

In a nutshell, our contribution can be concluded as:
\begin{itemize}
    \item We propose a MLLM-based dual momentum MoE \texttt{CL-MOE} framework for continual VQA, which not only alleviates the catastrophic forgetting issue but also improves the model's forward and backward transfer ability.
    \item To advance the MoE for CL, we design Dual-Router MoE (RMoE) and dynamic Momentum MoE (MMoE). First, the RMoE selects the most appropriate experts from local and global perspectives using instance-level and task-level routers. Then, the MMoE updates the experts dynamically based on the relationships among experts and tasks. 
    \item Through extensive experiments on VQA v2 datasets with 10 tasks, our \texttt{CL-MOE} achieves state-of-the-art performance for continual VQA by comparing with the strong baselines. 
\end{itemize}

\section{Related Work}
\label{sec:related}

\paragraph{Multimodal Large Language Models.} MLLMs~\cite{anil2023gemini,liu2024visual,wu2024next} refers to models based on LLMs~\cite{ouyang2022training,touvron2023llama2}, with the ability to receive, reason, and output multi-modal information. Since the release of GPT-4~\cite{achiam2023gpt}, there has been a fervent interest in researching MLLMs due to their impressive multi-modal task capabilities. Before MLLM, numerous efforts were dedicated to multimodal tasks, categorized into discriminative and generative paradigms, the representative works are CLIP~\cite{radford2021learning} and OFA~\cite{wang2022ofa}, respectively. 
The research on MLLM can be roughly divided into several categories: text and image~\cite{liu2024visual}, text and video~\cite{li2023videochat}, and text and audio~\cite{deshmukh2023pengi} content generation. 
However, most of these studies focus on learning the alignment and fusion among multiple modalities. In this paper, we apply MLLMs into a continual setting, to learn new knowledge without forgetting the history knowledge.

\paragraph{Continual Learning for LLMs.} 
In an era of rapid knowledge turnover, LLMs need to have the same mastery of knowledge as humans, retaining previously learned knowledge while absorbing new knowledge. However, LLMs exhibit catastrophic forgetting~\cite{kirkpatrick2017overcoming,ding2024boosting} when faced with a continuous data stream, leading to a decline in overall model generalization ability and degraded performance on previous tasks. Given the vast size of LLMs, retraining from scratch to incorporate new knowledge and instructions into the existing parts becomes impractical. 

Previously, many efforts have attempted to address the forgetting problem in MLLMs. 
CLAP4CLIP~\cite{jha2024clap4clip} utilizes a variational inference framework to probabilistically model the distribution of visually guided text features, improving fine-tuning reliability by considering the uncertainty in vision and text interactions. Adaptation-CLIP~\cite{liu2023class} introduces three different strategies for continuous learning of CLIP, namely linear adapter, self-attention adapter, and prompt tuning. 
Recently, VLM-PL~\cite{kim2024vlm} utilizes a visual-language model to optimize the pseudo-labeling process to improve the performance and accuracy of object detection models in CL scenarios. 
The most related study is VQACL~\cite{zhang2023vqacl}, it proposes a new continual learning framework for VQA tasks and introduces a novel representation learning strategy to enhance the model's reasoning and generalization ability. 
Most of these studies conduct experiments based on pre-trained models (e.g., T5, CLIP), which contain limited commonsense knowledge.
Unlike these studies, we learn task skills with multiple instance-level and task-level experts based on LLMs with huge parameters (e.g., LLaVA).   

\paragraph{Visual Question Answering.} 
VQA combines computer vision and natural language processing, aiming to enable models to answer natural language questions based on a given image. Recently, various methods~\cite{huai2023sqt,huai2024debiased} have been proposed for this task, and MLLMs have also demonstrated impressive performance~\cite{anil2023gemini,liu2024visual} in VQA tasks. However, these existing VQA models are trained offline, ignoring the requirement to handle continual multimodal data in practice. We apply continual learning to VQA and train the model with various tasks sequentially, which are more aligned with real-world non-stationary data streams.
\section{Preliminaries}
\label{sec:Preliminaries}
\subsection{Task Definition}
In this paper, we focus on continual visual question answering (VQA) tasks.
Unlike traditional offline training where the model has access to the entire training data, we concentrate on a continual learning setup in which the model accesses a non-stationary data stream. 
Specifically, we divide the data into $M$ subtasks based on question types, represented by the task descriptor $t \in \{1,2,\cdots, M\}$.
The $t_{th}$ subtask includes its specific training data $D_{t} = \{(I^{t}_{i}, O^{t}_{i})\}^{N_{t}}_{i=1}$ with $N_{t}$ tuples, where $I$, $O$ denotes the instruction (contains image and question) and output respectively. 
This task sequentially optimizes the MLLM on different VQA tasks, aiming to learn new knowledge of the current task while maintaining the knowledge of history tasks. 
In the test phase, we need to predict the label of examples from various tasks without knowing the task index.

\subsection{LoRA-based MoE}
Inspired by~\cite{dou2023loramoe,liu2023moelora,wu2024mixture,feng2024mixture,zhao2024retrieval}, we adopt Low-Rank Adaptation (LoRA) \cite{hulora2022} with a Mixture of Experts (MoE) \cite{jacobs1991adaptive,shazeer2016outrageously,du2022glam} framework. 
Specifically, MoE is a sparsely gated deep learning model that primarily consists of a set of experts and a router. 
The experts are several identical neural networks, and the router contains a gating function that models the probability distribution to generate the weights and weigh the outputs of these expert networks. 
The basic idea of MoE is to partition the input data into multiple partitions based on task class and assign data of each partition to one or more expert models. 
Each expert model can focus on processing specialized portions of the input data, thereby enhancing the overall performance of the model. 
The gating function receives intermediate representation $\mathbf{x}$ from the previous multi-head attention and outputs contributions to select the appropriate experts for the task, with weights generated by the following equation:
\begin{equation}
  G(\mathbf{x}) = \mathrm{Softmax}(\mathbf{x}\mathbf{W}_{gate}),
  \label{eq1}
\end{equation}
where $W_{gate}$ is the trainable weight in the gate function $G(\cdot)$, $\mathrm{Softmax}$ is used to normalize weights to balance the output distribution scale. Then, the output of the MoE layer can be expressed as:
\begin{equation}
   f(\mathbf{x}) = \sum^{n}_{i = 1}G(\mathbf{x})_{i} E_{i}(\mathbf{x}),
  \label{eq2}
\end{equation}
where $n$ is the number of experts, $E_{i}(\cdot)$ represent the output of $i_{th}$ expert and $G(\cdot)_{i}$ indicates $i_{th}$ value of the weight. 

In Transformer-based models, MoE usually replaces the feed-forward neural network (FFN) layer of each transformer block with an MoE layer. 
Considering model parameters and deploy cost, we adopt LoRA for MLLMs, freezing the original FFN layer parameters $\mathrm{W} \in \mathbb{R}^{in \times out}$ of the MLLM while replacing the experts' fully connected layers with low-rank matrices $\mathbf{A} \in \mathbb{R}^{in \times r}$ and $\mathbf{B} \in \mathbb{R}^{r \times out}$ to improve training efficiently:
\begin{equation}
   f(x) = \mathbf{W}\mathbf{x} + \frac{\alpha}{r} \sum^{n}_{i = 1}G(\mathbf{x})_{i} E_{i}(\mathbf{x}) = \mathbf{W}\mathbf{x} + \frac{\alpha}{r} \sum^{n}_{i = 1}G_{i}B_{i}A_{i}\mathbf{x},
  \label{eq3}
\end{equation}
where $\alpha$ and $r$ denote the constant hyper-parameter and rank, respectively. The matrices $A_{i} \in \mathbb{R}^{in \times \frac{r}{n}}$ and $B_{i} \in \mathbb{R}^{\frac{r}{n} \times out}$ indicate low rank matrices of the $i_{th}$ expert.
\begin{figure*}[!t]
\centering
\includegraphics[width=1\textwidth]{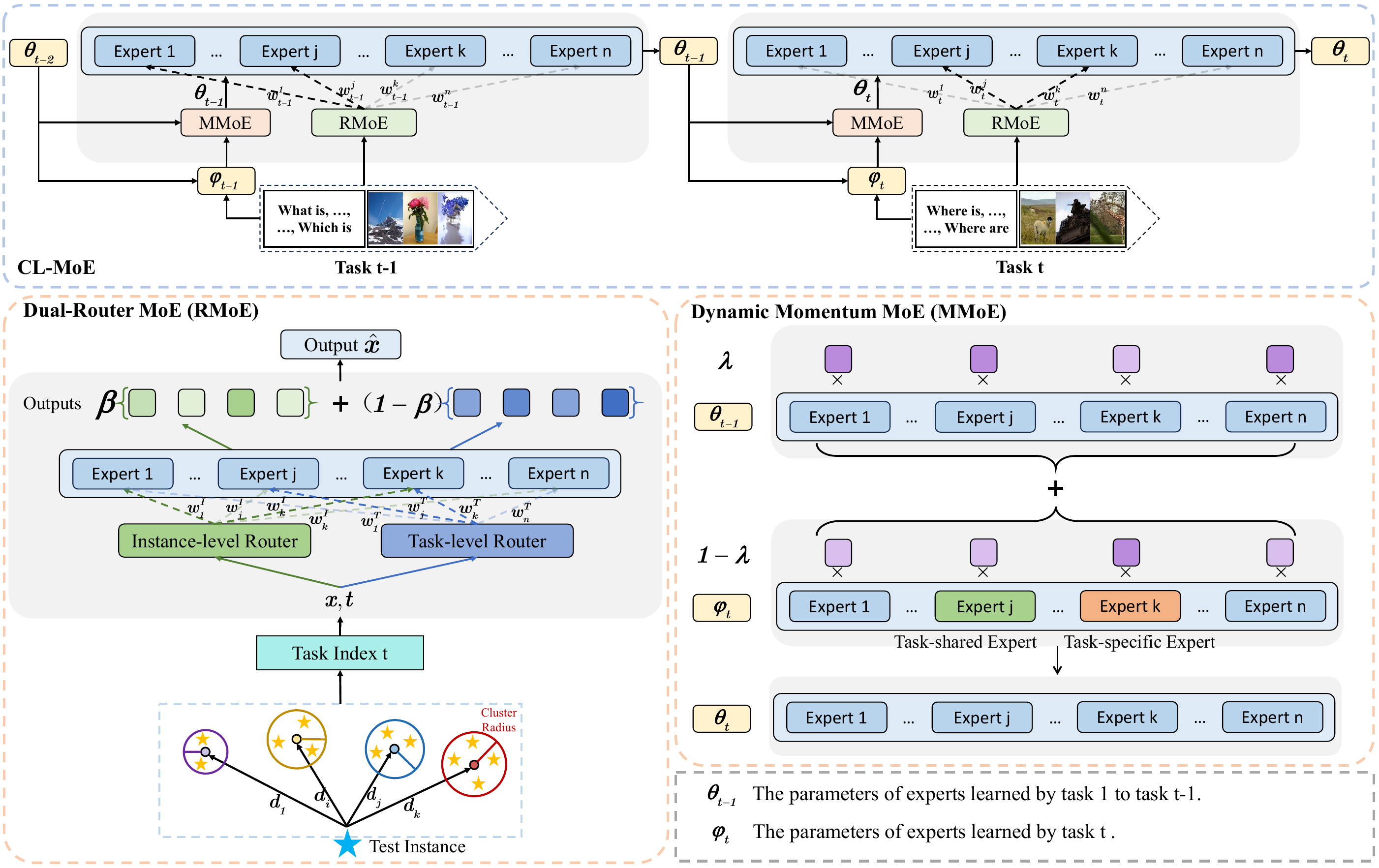}
\caption{The framework of our \texttt{CL-MoE} that contains Dual-Router MoE (RMoE) and Dynamic Momentum MoE (MMoE). 
We propose RMoE to capture the local and global experts using the instance-level and task-level routers. 
Then, MMoE dynamically updates the parameters of task-shared and task-specific experts selected by RMoE using a momentum strategy.
}
\label{framework}
\end{figure*}

\section{Method}
\label{sec:method}
In this section, we propose a dual momentum Mixture-of-Experts (\texttt{CL-MoE}) framework based on MLLMs for continual visual question answering, as shown in Figure~\ref{framework}. \texttt{CL-MoE} consists of two effective components, Dual-Router MoE (RMoE) and Dynamic Momentum MoE (MMoE). 
First, to select the most related experts, we design RMoE to capture the local and global experts using instance-level and task-level routers.
Then, we introduce MMoE, which updates the parameters of experts dynamically based on the experts selected by RMoE to retain useful knowledge and optimize irrelevant information.

\subsection{Overview}
In our study, we frame VQA as a generative task, intending to generate text answers from images and questions automatically. 
Before continual instruction tuning, the MLLM received abundant vision-language knowledge and instructions during the training phase to align the vision and language modalities.
Taking instruction $I$ as the input, which contains the image and question, MLLM calculates the probability of the entire output answer $O$ in an autoregressive way.
For example, one instruction template is `$<image>$ What is the person doing? Answer the question using a single word or phrase.' and the output $O$ is `skiing'. We optimize the model utilizing the following objective function:
\begin{equation}
  \mathcal{L} = -\sum^{L}_{j=1} \log p_{\Phi}(O_{j}|I, O_{1:j-1}),
  \label{eq4}
\end{equation}
where $L$ indicates the length of answer $O$, $O_{1:j-1}$ denotes all tokens before the index $j$ and $O_{j}$ means the $j_{th}$ token. $\Phi$ represents the trainable parameters of MLLMs.
Particularly, we adopt a LoRA-based MoE as the learnable parameters $\Phi=\{\theta, \Psi\}$, where $\theta = \{\theta^i\}_{i=1}^n$ is the parameters of experts and $\Psi = \mathbf{W}_{gate}$ is the parameters of the router. 
Here, $\theta^i = B_{i}A_{i}$ is the parameters of the $i_{th}$ expert $E_i$. 

In our setting, we train the model on a sequence of tasks in order. 
Let $\theta_{t-1}$ be the parameters of experts trained on task $\{1, 2, ..., {t-1}\}$ and $\varphi_t$ as the parameters of experts learned by $t$ task base on $\theta_{t-1}$. 
For each instance in the current task $t$, we obtain the intermediate representation $\mathbf{x}$ from the previous multi-head attention. Then, $\mathbf{x}$ is fed into the router and experts to generate weights and outputs, and the outputs of the experts are weighted for summation. 
For RMoE, we first train an instance-level router $G^I$ with experts' parameters $\varphi_t$ using dataset $D_t$. 
Then, we introduce the task-level router $G^T$ by calculating the average score of the weights output by the instance-level router. Then, we design MMoE to calculate $\theta_{t}$ based on $\theta_{t-1}$ and $\varphi_t$ using a dynamic momentum strategy. 
Specifically, we record the top $K$ experts that contribute the most to this task using task-level routing. 
We then split these experts into task-specific and task-shared experts and update their parameters dynamically.

\subsection{Dual-Router MoE}
In this paper, we assume one expert may serve more than one subtask (instance) and one subtask (instance) may require the collaboration of multiple experts. 
Furthermore, we believe that several fixed task-specific experts can solve most cases and some fine-grained experts should be considered according to the instance.
To address this problem, we present our Dual-Router MoE (RMoE) module, which aims to capture the local and global experts using instance-level and task-level routers. 

\paragraph{Instance-level Router.} In this module, we input the instance representation $\mathbf{x}$ into instance-level router $G_I$ to calculate the weights of the experts.
\begin{equation}
    \mathbf{w}^I = G^I(\mathbf{x}) = \mathrm{Softmax}(\mathbf{x}\mathbf{W}_{gate}) ,
    \label{eq5}
\end{equation}
where $\mathbf{w}^I = [w^I_1, ..., w^I_i, ..., w^I_n]$ is the weights of experts and $w^I_i$ is the weight of expert $E_i$.
We train the instance-level router on the training data $D_t$ so that the router learns to select the local experts based on the instance representation. Note that we also train the $\varphi_{t}$ in this step, which is initialized with $\theta_{t-1}$ and fine-tuned on $D_t$.

\paragraph{Task-level Router.} Unlike the instance-level router, we build a task-level router to capture the global experts $G^T(t)$, where $t$ is the task index. 
Specifically, we use the average weights of the instance-level router over the whole $D_t$ dataset to measure each expert's contribution to the $t$ task.
\begin{equation}
    \mathbf{w}^T = G^T(t) = \frac{1}{N^t} \sum_{\mathbf{x} \in D_t}G^I(\mathbf{x}).
    \label{eq6}
\end{equation}
During this period, we record the top $K$ (e.g., 2) experts that contribute the most to this task according to the weights output by $G^T(t)$, storing these task-level experts as ${\mathcal{E}_{t}}$. 

In the inference phase, since we do not know the task identifier, we present a task indexing module to obtain the task ID. 
First, we cluster the corpus of the training set according to the task descriptor and generate $M$ cluster centers. 
The cluster center $\mathbf{R}$ of $D_{t}$ can be expressed as,
\begin{equation}
  R_{t}=\frac{1}{|N_t|} \sum_{x \in D_{t}} F(x) ,
  \label{eq7}
\end{equation}
where $F(x)$ represents the text feature of sample $x$, which is the hidden state of the $CLS$ token after the MLLM encoder.
We determine the task identifier of test instance $v$ by finding the nearest anchor $R_{t}$.
\begin{equation}
  t = \arg\min_{t \in \{1, 2, \ldots, M\}} \| F(v) - R_t \|_2 ,
  \label{eq8}
\end{equation}
where $\|\cdot\|_{2}$ denotes the Euclidean distance. 

Finally, we utilize instance-level and task-level weights to obtain the comprehensive representation $\hat{\mathbf{x}}$ based on the task descriptor $t$ and the intermediate representation $\mathbf{x}$,
\begin{equation}
\begin{aligned}
    \hat{\mathbf{x}} & = \beta \frac{\alpha}{r} \sum^{n}_{i = 1}G^I(\mathbf{x})_{i} E_{i}(\mathbf{x}) + (1-\beta) \frac{\alpha}{r}\sum^{n}_{i = 1} G^T(t)_{i} E_{i}(\mathbf{x}) \\
     & = \beta\sum^{n}_{i = 1}w^I_{i}\theta^t_i\mathbf{x} + (1-\beta)\sum^{n}_{i = 1}w^T_{i}\theta^{t}_i\mathbf{x} ,  
\end{aligned}
\label{eq9}
\end{equation}
where $\beta$ is a hyper-parameter to balance the local (instance-level) and global (task-level) weights.

\begin{table*}[!t]
\centering
\setlength{\tabcolsep}{1.4mm}{
\begin{tabular}{lcccccccccccc}
\hlineB{4}
\multirow{2}{*}{Methods} & \multicolumn{10}{c}{Various task in VQA v2}                                                                                                                             & \multirow{2}{*}{$AP(\uparrow)$} & \multicolumn{1}{l}{\multirow{2}{*}{$AF(\downarrow)$}} \\ \cline{2-11}
                         & Rec.           & Loc.           & Jud.           & Com.           & Cou.           & Act.           & Col.           & Typ.           & Sub.           & Cau.           &                     & \multicolumn{1}{l}{}                     \\ \hline
\multicolumn{13}{l}{\emph{VL-T5 based methods}}                                                                                                                                                                                                        \\ \hline
Vanilla                  & 7.39           & 4.94           & 22.29          & 32.30          & 0.71           & 12.14          & 12.10          & 10.69          & 27.29          & 15.10          & 14.49               & 30.15                                    \\
EWC                      & 6.73           & 8.43           & 27.22          & 47.10          & 0.14           & 12.40          & 1.76           & 10.98          & 31.05          & 11.85          & 15.77               & 28.38                                    \\
MAS                      & 30.81          & 8.07           & 25.50          & 4.00           & 31.90          & 32.39          & 26.24          & 24.75          & 19.85          & 2.75           & 20.56               & 21.97                                    \\
ER                       & 18.64          & 21.36          & 61.27          & 64.17          & 30.29          & 52.84          & 43.39          & 23.31          & 42.75          & 11.85          & 36.99               & 4.80                                     \\
DER                      & 14.55          & 13.83          & 62.88          & 65.16          & 30.96          & 51.19          & 40.51          & 19.04          & 42.87          & 12.55          & 35.35               & 6.58                                     \\
VS                       & 15.66          & 19.21          & 59.86          & 32.16          & 27.28          & 47.79          & 32.32          & 20.44          & 41.38          & 10.20          & 34.03               & 11.68                                    \\
VQACL                    & 20.47          & 28.02          & 62.55          & 68.61          & 29.35          & 50.66          & 44.45          & 26.36          & 44.65          & 12.60          & 38.77               & 2.90                                     \\
Multitask                    & 42.89          & 38.27          & 75.96          & 73.34          & 38.01          & 66.90         & 56.52          & 47.46       & 53.59          & 22.94         &  -              & -                                        \\
\hline
\multicolumn{13}{l}{\emph{LLaVA-7B based methods}}                                                                                                                                                                                                     \\ \hline
Vanilla                  & 19.25          & 14.81          & 54.59          & 56.97          & 24.23          & 46.20          & 27.58          & 26.09          & 36.47          & 18.89          & 32.51               & 20.69                                    \\
EWC                      & 28.12          & 23.02          & 61.50          & 61.08          & 26.13          & 54.29          & 23.65          & 32.25          & 44.97          & 17.83          & 37.28               & 15.27                                    \\
MAS                      & 31.54          & 22.09          & 60.85          & 46.32          & 32.48          & 56.47          & 30.05          & 35.69          & 42.73          & 18.83          & 37.71               & 14.91                                    \\
ER                       & 29.31          & 25.74          & 63.46          & 65.78          & 31.92          & 58.39          & \textbf{45.17}          & 34.55          & 46.24          & 18.96          & 41.95               & 10.20                                    \\
DER                      & 26.95          & 21.43          & 64.88          & 66.17          & 31.01          & 55.92          & 44.60          & 32.85          & 47.09          & 20.74          & 41.16               & 11.28                                    \\
VS                       & 28.48          & 24.09          & 61.37          & 67.20          & 29.56          & 54.64          & 33.98          & 32.91          & 45.82          & 19.89          & 39.79               & 12.70                                    \\
VQACL                       & 34.14          & 32.19          & 66.15          & 63.00          & 33.01          & 60.91          & 34.64          & 38.48          & 47.94          & \textbf{24.42}          & 43.49               & 9.10                                    \\ \hline
\texttt{CL-MoE}           & \textbf{46.50} & \textbf{37.18} & \textbf{75.22} & \textbf{71.39} & \textbf{40.90} & \textbf{69.54} & 43.66 & \textbf{52.68} & \textbf{55.55} & 20.74 & \textbf{51.34}      & \textbf{-0.02}                                     \\
Multitask (Upper Bound)                    & 55.15          & 41.88          & 80.74         & 75.47         & 49.81          & 75.97         & 73.03         & 61.02       & 60.54          & 29.49          & -              & -                                        \\
\hlineB{4}
\end{tabular}}
\caption{Performance (\%) of our \texttt{CL-MoE} and distinct continual learning method on VQA v2. We list the accuracy for each task along with $AP$ and $AF$. The best results are emphasized in \textbf{bold}.}
\label{tab1}
\end{table*}

\subsection{Dynamic Momentum MoE}
An ideal MLLM in a continual setting needs to be equipped with knowledge retention capabilities and able to use the recently learned knowledge to solve previous and subsequent tasks, \ie backward transfer and forward transfer ability. 
We introduce the Dynamic Momentum MoE (MMoE) to enhance its anti-forgetting and transfer capabilities. 
The $\varphi_{t}$ is initialized by $\theta_{t-1}$ and tuning based on the dataset $D_t$, which contains rich knowledge of the current task and $\theta_{t-1}$ contains the knowledge of history tasks from task $1$ to $t-1$. 
To control the balance of $\varphi_{t}$ and $\theta_{t-1}$, we propose a momentum strategy to update the parameters $\theta_t$ dynamically. 

Based on the task-level experts $\mathcal{E}_t$ and $\mathcal{E}_{pre}=\mathcal{E}_{1} \cup ... \cup \mathcal{E}_{t-1}$ selected by RMoE, we split all the experts into task-shared experts, task-specific experts, and none. 
(1) Task-shared experts mean the expert occurs in $\mathcal{E}_t$ and $\mathcal{E}_{pre}$ at the same time. This indicates that the expert contributes significantly to both the $t_{th}$ task and previous tasks. We consider task $t$ and previous tasks require similar skills, primarily retaining parameter $\theta_{t-1}$; 
(2) task-specific experts mean the expert only occurs in $\mathcal{E}_t$ and does not occur in $\mathcal{E}_{pre}$. It indicates that the expert's ability significantly contributes to task ${t}$ but less to previous tasks. Thus we primarily retain parameter $\varphi_{t}$ during subsequent dynamic momentum updates; 
and (3) none means the expert does not occur in $\mathcal{E}_t$. This indicates that the expert has no remarkable contribution to the current task, so we mainly keep the parameters $\theta_{t-1}$.
The above progress can be summarized as:
\begin{equation}
  \lambda_{i} = 
    \begin{cases} 
    \gamma, & \ { \text{if}\ E_{i}\ \text{is task-shared expert}} \\
    1 - \gamma, &  \ { \text{if}\ E_{i}\ \text{is task-specific expert}} \\
    \gamma, &  \ { \text{otherwise}}  \\
    \end{cases}
    \label{eq10}
\end{equation}
where $\lambda_i$ is the weight for expert $E_i$. Here $\gamma$ is a hyper-parameter, where $\gamma > 0.5$. Finally, we obtain the weight vector for all experts,
\begin{equation}
    \mathbf{\lambda} = [\lambda_{1},\cdot \cdot,\lambda_{i},\cdot \cdot,\lambda_{n}].
\end{equation}
We then perform dynamic momentum updates parameters of experts based on $\theta_{t-1}$ and $\varphi_{t}$ using the vector $\mathbf{\lambda}$, as shown follows:
\begin{equation}
  \theta_{t} = \mathbf{\lambda} \cdot \theta_{t-1} + (\mathbf{1}-\mathbf{\lambda}) \cdot \varphi_{t},
  \label{eq12}
\end{equation}
where $\theta_{t}$ represents the updated expert parameters for task ${t}$. The $\cdot$ and $+$ indicate element-wise multiplication and addition operations. By incorporating MMoE, we can integrate new knowledge while preserving old knowledge effectively, thus not only mitigating catastrophic forgetting but also boosting the backward transfer ability of the model.

\section{Experimental Setups}
\label{sec:exp}

\paragraph{Dataset and Evaluation Metrics.}

We conduct experiments on the VQA v2~\cite{goyal2017making} benchmark, which includes over 200k images and 1.1M questions, where images are primarily from the COCO~\cite{lin2014microsoft} dataset. Following the VQACL setup~\cite{zhang2023vqacl}, we divided the VQA v2 into 10 tasks based on question types: recognition, location, judge, commonsense, count, action, color, type, subcategory, and causal. 

We use two standard continual learning evaluation metrics~\cite{chaudhry2018riemannian,lopez2017gradient} to measure the performance of \texttt{CL-MoE}: Final Average Performance ($AP$) and Average Forgotten ($AF$). Specifically, $AP$ is the average performance on all tasks after the continual fine-tuning ends, reflecting the model's ability to maintain learned knowledge while learning new knowledge. Let $m_{a,b}$ denote the test performance on task ${b}$ upon completing the training of task ${a}$, $AP = \frac{1}{M} \sum^{M}_{i=1} m_{M,i}$, where $M$ denotes the number of tasks. $AF$ represents the performance on previous tasks after learning new tasks compared to the fine-tuning performance on old tasks, which also reflects the average forgetting on past tasks. $AF = \frac{1}{M-1} \sum^{M-1}_{i=1} m_{i,i} - m_{M,i}$. According to~\cite{cho2021unifying}, we use the accuracy percentage as the $m$.

\begin{table*}[!t]
\centering
\setlength{\tabcolsep}{1.6mm}{
\begin{tabular}{l|cc|cccccccccc|cc}
\hlineB{4}
  & \multicolumn{2}{c|}{Method} & \multicolumn{10}{c|}{Various task in VQA v2}                                       & \multirow{2}{*}{AP} & \multirow{2}{*}{AF} \\ \cline{1-13}
  & MMoE      & RMoE      & Rec.  & Loc.  & Jud.  & Com.  & Cou.  & Act.  & Col.  & Typ.  & Sub.  & Cau.  &                     &                     \\ \hline
a & $\times$     & $\times$     & 19.25 & 14.81 & 54.59 & 56.97 & 24.23 & 46.20 & 27.58 & 26.09 & 36.47 & 18.89 & 32.51 
   & 20.69               \\
b & $\surd$      & $\times$     & 42.84 & 34.59 & 72.11 & 69.30 & 36.76 & 65.62 & 39.95 & 50.02 & 53.98 & 19.11 & 48.43               & 3.02                \\
c & $\times$     & $\surd$      & 27.36 & 25.62 & 64.01 & 65.96 & 31.52 & 56.23 & 37.17 & 38.26 & 46.49 & 19.70 & 41.23               & 11.09               \\
d & $\surd$      & $\surd$      & \textbf{46.50} & \textbf{37.18} & \textbf{75.22} & \textbf{71.39} & \textbf{40.90} & \textbf{69.54} & \textbf{43.66} & \textbf{52.68} & \textbf{55.55} & \textbf{20.74} & \textbf{51.34}      & \textbf{-0.02}                \\ \hlineB{4}
\end{tabular}
}
\caption{Ablation study of our \texttt{{CL-MoE}} on VQA v2.}
\label{tab:ac2}
\end{table*}

\paragraph{Baselines.}
To demonstrate the effectiveness of our method, we select several typical continual learning methods, including replay-based methods and regularization-based methods. For replay-based methods, we adopt ER~\cite{chaudhry2019continual}, DER~\cite{buzzega2020dark}, VS~\cite{wan2022continual} and VQACL~\cite{zhang2023vqacl}. For regularization-based methods, we compare with EWC~\cite{kirkpatrick2017overcoming} and MAS~\cite{aljundi2018memory}. 
Multitask represents the performance of the model that trains on all the tasks once, while Vanilla indicates the performance of the model trained on a sequence of tasks without using any additional methods. Please find more details about the baselines in the Appendix.

For a fair comparison, we conduct the experiments on both VL-T5~\cite{cho2021unifying} and LLaVA~\cite{liu2024visual}. 
Particularly, VL-T5 is a unified framework that extends the pre-trained language model T5~\cite{raffel2020exploring} with visual understanding capabilities. 
LLaVA-7B~\cite{liu2024visual} is a MLLMs-based model connecting the open-set visual encoder of CLIP~\cite{radford2021learning} and Vicuna~\cite{chiang2023vicuna}. It is fine-tuned on the visual language instruction-following dataset, which includes three types of instruction-following data: conversational QA, detailed descriptions, and complex reasoning.

\paragraph{Implementation Details.}
In the experiments, we use LLaVA-7B as our MLLM for continual tuning. It employs Vicuna~\cite{chiang2023vicuna} as LLM and pre-trained CLIP visual encoder ViT-L/14~\cite{liu2021swin} to extract visual embeddings from images of size 336 $\times$ 336px. We set the embedding dimension to 64. For the rehearsal method, we set the memory bank size to 5000. For our proposed MMoE and RMoE, we configure the number of experts $n$ to 8, record top expert $K$ to 2, the rank $r$ to 64, the hyperparameter $\alpha$ to 128, $\gamma$ to 0.7, and $\beta$ to 0.5. During training, we train each task for 1 epoch with a batch size of 16. We use the AdamW as the optimizer with the learning rate of 2$e^{-4}$, and employ the cosine learning rate scheduler.

\section{Experimental Analysis}

\subsection{Main Results}
We report the experimental results of baselines and our \texttt{CL-MoE} over 10 tasks, as shown in Table~\ref{tab1}.
From the results, we obtain the following findings. 
\textbf{First}, our method achieves state-of-the-art performance in most cases by comparing with both VL-T5 and LLaVA. For example, our model outperforms the previous SOTA baseline VQACL in terms of $AP$ and $AF$.
\textbf{Second}, compared to Vanilla LLaVA, our \texttt{CL-MoE} improved $AP$ by approximately 14.36\% (51.34\% vs. 36.98\%), with substantial performance gains across all tasks. For $AF$, \texttt{CL-MoE} improves the performance by approximately 20.71\% (-0.02\% vs. 20.69\%). It is worth noting that our $AF$ value is less than 0, which means our average performance on the 9 tasks is even better than the fine-tuning performance, proving that our method has favorable backward transfer capability. In addition, our method also outperforms fine-tuning on the last task, proving that our method has excellent forward transfer ability. These observations show that our model not only improves the average accuracy but also significantly mitigates the forgetting problem. \textbf{Third}, Methods based on LLaVA-7B generally achieve better average accuracy compared to those based on VL-T5, indicating that larger models can better exploit the potential of learning multimodal knowledge, making them more suitable for visual question answering. However, it is worth noting that the $AF$ performance of rehearsal-based methods on LLaVA is worse than on VL-T5, whereas regularization-based methods showed the opposite trend. We believe that while larger MLLMs can improve performance, they are also more susceptible to the forgetting problem. 
\textbf{Fourth}, compared with the upper bound method Multitask that trains on the merged datasets of all the subtasks, our model still has room to improve. We would like to explore a more effective algorithm for continual multi-modal tasks in the future.

\subsection{Ablation Study}
To investigate the effectiveness of each component in our method, we conduct ablation experiments on \texttt{CL-MoE}, and the results are shown in Table 2. Specifically, we conduct experiments with only MMoE, only RMoE, and the complete components. 
By comparing (a, b) and (a, c), we can conclude that both modules we designed contribute to the continual tuning based on MLLM. To be specific, MMoE updates the expert parameters based on the designed momentum strategies, allowing experts to integrate new knowledge and instructions while retaining the original knowledge. MMoE plays an important role in \texttt{CL-MoE}. For RMoE, it robustly allocates the most suitable experts for solving the problem, integrating instance-level and task-level routing. It is worth noting that using RMoE alone does not achieve outstanding performance, because a considerable amount of knowledge is lost during the training phase without MMoE. 
Even if the most suitable experts are allocated, the selected experts might lose part of their capability to solve the problem. 
By comparing (a, d), we conclude that our two components work closely together, effectively mitigating the forgetting problem and improving the transfer abilities in continually fine-tuning MLLM for VQA.

\begin{figure*}[t!]
    \centering
    \begin{minipage}[b]{0.48\textwidth}
        \centering
        \includegraphics[width=\linewidth]{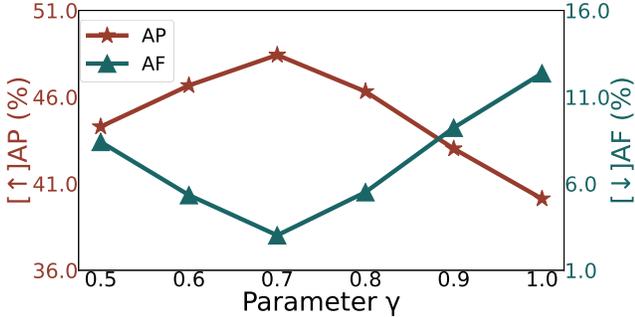}
    \end{minipage}
    \hfill
    \begin{minipage}[b]{0.48\textwidth}
        \centering
        \includegraphics[width=\linewidth]{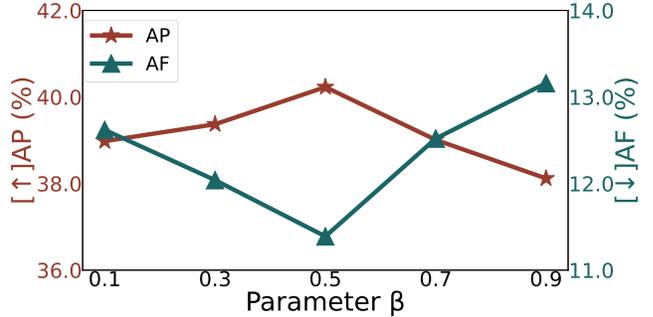}
    \end{minipage}
    \caption{Performance(\%) of our \texttt{{CL-MoE}} with different hyperparameters $\gamma$ and $\beta$ on VQA v2.}
    \label{fig:hyperparameters}
    \vspace{-0.3em}
\end{figure*}

\subsection{Further Analysis}
\paragraph{Impact of Hyperparameter $\gamma$.}  We investigate the impact of critical hyperparameters used in our method, specifically $\gamma$ in MMoE, as illustrated in Figure~\ref{fig:hyperparameters}. It is used to control the balance between the current task parameters $\theta_{t}$ and the previous task parameters $\theta_{t-1}$ during momentum updates in MMoE. 
Since our configuration sets $\gamma > 0.5$, we assign $\gamma \in \{0.5, 0.6, 0.7, 0.8, 0.9, 1.0\}$. 
Small $\gamma$ leads to the \texttt{CL-MoE} model focus on the history task while the new knowledge of the current task is not transferred.
The model with large $\gamma$ captures the knowledge of the current task while forgetting the abilities of history tasks.
The results indicate that the best performance is achieved when $\gamma = 0.7$. 
In this setting, \texttt{CL-MoE} retains most of the knowledge while absorbing new knowledge. 

\vspace{-1em}
\paragraph{Impact of Hyperparameter $\beta$.} We investigate the impact of hyperparameters $\beta$ in RMoE. It balances the experts and their weights chosen by the Task-level Router and Instance-level Router in RMoE. 
We configure $\beta \in \{0.1, 0.3, 0.5, 0.7, 0.9\}$, as represented in Figure~\ref{fig:hyperparameters}. 
The results show that the best performance is achieved when $\beta = 0.5$. We discover performance drops when $\beta$ is extremely large or small, it implies that both instance-level and task-level routers are important for RMoE. We follow this setup in subsequent experiments to seek a balance.
\vspace{-1em}

\begin{table}[!t]
\small
\centering
\begin{tabular}{c|ccc|ccc}
\hlineB{2}
\multirow{1}{*}{Method} & \multirow{1}{*}{$n$} & \multirow{1}{*}{AP} & \multirow{1}{*}{AF} & \multirow{1}{*}{$K$} & \multirow{1}{*}{AP} & \multirow{1}{*}{AF} \\
\hline
\multirow{4}{*}{\texttt{CL-MoE}} & 1                               & 32.51               & 20.69             & 1 & 50.64 & 0.69  \\
                        & 2                               & 44.12               & 8.40               & 2 & 51.34 & -0.02    \\
                        & 4                               & 50.19               & 1.15              & 3 & 51.22 & 0.30     \\
                        & 8                               & 51.34               & -0.02               & 4 & 50.93 & 0.17    \\ \hlineB{2}
\end{tabular}
\caption{Performance(\%) of our \texttt{{CL-MoE}} with various experts number $n$ and top $K$ experts with $n=8$ on VQA v2.}
\label{tab:ac3}
\vspace{-0.5em}
\end{table}

\paragraph{Impact of Number of Experts $n$ and Top $K$ Experts.} We study the impact of the number of experts $n$ and top $K$ experts for each task on our \texttt{CL-MoE}, as shown in Table~\ref{tab:ac3}. The experimental results show that our method achieves sub-optimal performance with 4 experts and reaches the optimal level when $n = 8$. This means we can effectively address the forgetting issue in MLLM using MMoE and RMoE with minimal resource overhead. When the number of experts is few, there is a significant drop in performance. We think that MMoE and RMoE cannot fully leverage their advantages with too few experts. 

Furthermore, we increase the top $K$ from 1 to 4 under $n=8$. From the results, we observe that two task-specific experts are optimal for our proposed \texttt{CL-MoE}. We consider excessive task-specific experts to be redundant, whereas few task-specific experts are insufficient to effectively address the task. Moreover, the performance differences when $K$ is assigned different values are not significant, this demonstrates the robustness of our method. In our experimental setup, we set $K$ to 2 to achieve a satisfactory trade-off between resources and performance. Please infer the supplementary material for complete experimental results.

\vspace{-1em}
\paragraph{Impact of the Order of Tasks.} We investigate the impact of different task orders on our \texttt{CL-MoE}. Specifically, we use the reverse order of the original setting for continual tuning on VQA v2, as shown in Table \ref{tab:ac4}. The experimental results indicate that \texttt{CL-MoE} also achieves optimal performance on the new task order. Our model outperforms the SOTA methods VQACL by more than 6 points in terms of AP (57.08 vs. 50.73). 
Additionally, we find that the task order has a significant impact on performance. 
We observe that the reverse task order performs better than the forward order (57.08 vs. 51.34 and -1.44 vs. -0.02). Due to the different task correlations, the task order will influence the difficulties of forgetting and transferring during the learning process. Please infer the supplementary material for complete experimental results.

\begin{table}[!t]
\centering
\begin{tabular}{l|cc|cc}
\hlineB{2}
\multirow{2}{*}{Method}      & \multicolumn{2}{c|}{Forward}                       & \multicolumn{2}{c}{Reverse} \\ 
                             & \multicolumn{1}{c}{AP} & AF                        & AP           & AF           \\ \hline
\texttt{CL-MoE} & 51.34                  & \multicolumn{1}{l|}{-0.02} & 57.08        & -1.44        \\
VQACL                        & 43.49                  & \multicolumn{1}{l|}{9.10} & 50.73        & 4.91         \\
\hlineB{2}
\end{tabular}
\caption{Performance(\%) of our \texttt{{CL-MoE}} with reverse task order on VQA v2.}
\label{tab:ac4}
\end{table}

\section{Conclusions and Further Works}
\label{sec:conclusion}

In this paper, we propose the \texttt{CL-MoE} framework on instruction tuning MLLM for continual VQA tasks. To appropriately assign experts, we introduce RMoE which contains the instance-level and task-level routers, from local and global perspectives to robustly allocate weights to the corresponding experts. To alleviate the forgetting problem and improve the transfer capabilities of the model, we designed MMoE to update the parameters of task-specific and task-shared experts using a dynamic momentum update strategy. Extensive experiments on VQA v2 demonstrate that our method achieves optimal performance by comparing with previous SOTA baselines, proving its anti-forgetting and transfer capabilities. Ablation studies also confirm the effectiveness of the \texttt{CL-MoE}'s components. In the future, we aim to extend continual learning-based MLLMs to other diverse tasks, further addressing the forgetting problem in MLLMs for continual multitask learning.

\section*{Acknowledge}
The authors wish to thank the reviewers for their helpful comments and suggestions.
This research is funded by the National Science and Technology Major Project (No. 2021ZD0114002), the National Nature Science Foundation of China (No. 62477010 and No. 62307028), the Natural Science Foundation of Shanghai (No. 23ZR1441800), Shanghai Science and Technology Innovation Action Plan (No. 24YF2710100 and No. 23YF1426100 ) and Shanghai Special Project to Promote High-quality Industrial Development (No. 2024-GZL-RGZN-02008).

{
    \small
    \bibliographystyle{ieeenat_fullname}
    \bibliography{main}
}


\end{document}